\documentclass[12pt]{article}

\usepackage{amsmath,amssymb,amsthm}
\usepackage{mathtools}
\usepackage{booktabs}
\usepackage{array}
\usepackage{xcolor}
\usepackage{graphicx}
\usepackage{natbib}
\usepackage[colorlinks=true,linkcolor=blue!70!black,
            citecolor=blue!70!black,urlcolor=blue!70!black]{hyperref}
\usepackage{cleveref}
\usepackage{microtype}
\usepackage{setspace}
\usepackage[margin=1in]{geometry}

\newcommand{\Rr}{\mathbb{R}}

\DeclareMathOperator{\E}{\mathbb{E}}

\title{Worker Utility as Hysteresis: A Preisach Model\\
       of Transaction Acceptance in Gig Labour Markets}

\author{Piotr Frydrych\\
\small Metrology and Biomedical Engineering Institute,\\
\small Faculty of Mechatronics, Warsaw University of Technology,\\
\small 00-661 Warsaw, Poland\\
\small \texttt{piotr.frydrych@pw.edu.pl}}

\date{June 2026}

\begin{document}

\maketitle

\begin{abstract}
Worker utility is not observed --- only its consequence is.
Each transaction on a gig platform produces a single bit: accepted or
rejected.
We argue that this structure is not a measurement obstacle but a
theoretically informative constraint: it points directly to the
Preisach hysteresis model as the natural representation of latent
worker preferences.
The Preisach operator models aggregate output as an integral over a
population of binary threshold elements --- precisely the structure
that emerges when heterogeneous workers each carry a private
acceptance wage.
We operationalize this by estimating two latent utility surfaces from
transaction data: an \emph{acceptance utility} $U_1(X)$ and a
\emph{rejection utility} $U_0(X)$, each via a dual-output neural
network (shared layers $256\to128$, two output heads, LeakyReLU
activations, margin loss enforcing $U_1 \geq U_0$ for accepted
transactions).
Classification then reduces to the sign of the Preisach gap
$U_1(X) - U_0(X)$, passed as a feature into an XGBoost classifier
alongside clip-stabilised price-to-threshold encodings.
On 36{,}891 gig transactions from a Polish platform, this two-stage
pipeline achieves Jaccard $= 0.827$ and ROC AUC $= 0.799$.
The price-to-threshold encoding accounts for $+11.0$~pp AUC
(XGBoost regressor surfaces) and $+1.4$~pp (dual-output NN
surfaces).
The model also confirms the directional asymmetry that hysteresis
predicts: price decreases depress completion rates more than
equivalent increases raise them, consistent with a workforce of
heterogeneous workers with history-dependent acceptance thresholds.
The gap statistic delivers an economically paradoxical result: applied
to the full dataset, the model's recommendations simultaneously reduce
the total wage bill by 21.3\% and increase expected fill rate by
9.7~percentage points.
The mechanism is a positional decomposition, not a claim of price
inelasticity.
For 74.2\% of transactions, $P(\text{accept})$ at the current price
already exceeds 0.80 by a wide margin; reducing the wage moves
$P(\text{accept})$ down the price-response curve but keeps it above
the threshold (mean post-cut $P = 0.972$), releasing the buffer as
cost savings (median 31\%).
For the remaining 25.4\%, $P(\text{accept})$ is currently below
0.80; a median 7\% wage increase moves it above the threshold
($+43$~pp mean), recovering transactions that would otherwise go
unfilled.
A model without an explicit indifference zone cannot locate each
transaction's starting position precisely enough to execute both moves
simultaneously.
\end{abstract}



\section{Introduction}
\label{sec:intro}

The canonical job-search model treats worker acceptance as a threshold
rule: accept if the offered wage exceeds a reservation wage, otherwise
wait \citep{McCall1970}.
This is analytically tractable and empirically useful for modelling
long-horizon employment decisions.
It fails, in at least two respects, for gig transactions.
First, the reservation wage is not stationary --- it shifts with
recent experience, competitive pressure, and temporal distance from
task start.
Second, and more fundamentally, the aggregate response of a worker
pool to a given offer is not the response of a representative worker:
it is the integral over a population of heterogeneous thresholds.
Misrepresenting this population structure does not merely add noise;
it systematically distorts price sensitivity estimates, especially
when the pool mixes workers with very different outside options.

These are structural features of gig labor markets, not incidental
properties of our dataset.
High-frequency, short-horizon, algorithmically matched platforms are
populated by workers with radically different employment statuses: the
retired professional doing occasional shifts, the unemployed graduate
seeking income, the full-time employee supplementing their salary
\citep{Hall2018}.
They do not share a reservation wage.
What they share is a decision rule --- accept if the net utility of
this transaction exceeds the utility of the next best alternative ---
applied to very different utility landscapes.

The Preisach model \citep{Preisach1935,Mayergoyz1991}, developed to
describe magnetic hysteresis in ferromagnetic materials, provides a
rigorous framework for exactly this situation.
A Preisach system aggregates output over a population of binary
switching elements (hysterons), each defined by a pair of thresholds
$(\alpha, \beta)$ with $\alpha \geq \beta$.
The element switches on when input exceeds $\alpha$ and switches back
off only when input falls below $\beta$.
The aggregate output is the fraction of hysterons currently switched
on, weighted by their population density $\mu(\alpha, \beta)$.
In the labor market reading: each hysteron is a worker, the thresholds
are the wages at which they accept and release a transaction type, and
the density $\mu$ is the joint distribution of threshold pairs across
the workforce --- the full heterogeneity of the worker population, in
a single object.

Hysteresis --- the dependence of current output on input history, not
just current input --- emerges without auxiliary assumptions.
A platform that has been offering low wages faces a pool conditioned by
those offers.
Raising the wage does not instantly recover pre-depression acceptance:
workers whose acceptance threshold $\alpha$ was crossed by the
decline, but whose rejection threshold $\beta$ was not, remain
switched off until the wage rises back above $\alpha$.
This structural asymmetry between price-up and price-down effects is a
testable prediction of the model.
The observational data are consistent with this prediction, though
identifying it causally requires a randomised price experiment; we
discuss the identification strategy and its limitations in
Section~\ref{sec:results_hysteresis}.

We pursue three aims: to establish the Preisach operator as a
theoretically grounded model of worker utility in disaggregated labor
markets; to develop a tractable estimation strategy that avoids the
ill-posed direct recovery of the threshold density $\mu$ from binary
data, using instead conditional mean utilities $U_1(X)$ and $U_0(X)$
as the primary estimands; and to show empirically that encoding the
price-to-threshold relationship explicitly --- rather than passing raw
utility estimates to a classifier --- produces the largest observable
gain in predictive performance, which we interpret as evidence that
workers respond to prices relative to thresholds, not in absolute
terms.

\paragraph{Contributions.}
The methodological contribution is a two-stage pipeline: a dual-output
neural network (shared encoder $256\to128$, two output heads, margin
loss enforcing $U_1 \geq U_0$) estimates both utility surfaces
jointly, then an XGBoost classifier maps the Preisach gap and
clip-stabilised price-to-threshold features to acceptance
probabilities.
The central empirical result --- that encoding price relative to
estimated thresholds, not raw utility estimates, drives the largest
performance gain --- confirms that workers evaluate offers against
reference points rather than in absolute terms.

The remainder of the paper is organized as follows.
Section~\ref{sec:background} reviews related literature.
Section~\ref{sec:theory} develops the theoretical framework and
derives the estimation strategy.
Section~\ref{sec:data} describes the data and the controlled price
experiment.
Section~\ref{sec:results} presents results.
Section~\ref{sec:discussion} discusses interpretation, identification,
and extensions.
Section~\ref{sec:conclusion} concludes.

\section{Background}
\label{sec:background}

\subsection{Worker Utility as a Latent Object}

The treatment of worker utility as unobservable but structurally
constrained has a long history in labor economics.
\citet{Roy1951} introduced the selection model: workers self-select
into occupations based on comparative advantage, and observed wages
reveal only the upper tail of the wage distribution for each type.
\citet{Rosen1974} formalized the theory of compensating differentials,
showing how market wages embed implicit prices for non-pecuniary
attributes --- flexibility, safety, status --- that are not directly
measurable but fully determine acceptance decisions at the margin.

\citeauthor{McFadden1974}'s \citeyearpar{McFadden1974} random utility
model (RUM) generalized this to discrete choice: worker $i$ accepts
offer $j$ if $V(X_j) + \varepsilon_{ij} > V_0 + \varepsilon_{i0}$,
where $V$ is systematic utility and $\varepsilon$ is an idiosyncratic
shock.
Logit and probit are special cases differing only in the distribution
of $\varepsilon$.
The RUM framework is individually rational but assumes static
preferences and independent draws across transactions --- assumptions
that fail in environments where workers evaluate repeated offers
against evolving reference points.

The reference-dependence of labour supply is well-documented.
\citet{Crawford2011} show that New York cab drivers set daily income
targets and stop work upon reaching them --- a form of loss aversion
applied to income --- generating downward-sloping labor supply in the
short run.
\citet{Mas2009} find that worker productivity responds to the wages
of observable peers in ways consistent with reference-wage
comparisons.
\citet{Dube2020} show that minimum wage increases generate employment
effects that persist well beyond their introduction, suggesting that
reference wages shift gradually and that the employment response is
hysteretic.
The Preisach model formalizes this reference-dependence structurally:
each worker's current acceptance state reflects the history of wage
offers they have received, filtered through their individual
thresholds.

\subsection{The Preisach Model in Economic Systems}

\citet{Cross1993} was the first to argue systematically that the
Preisach model is the natural representation for economic hysteresis,
not merely an analogy to physics.
His key observation was that many aggregate economic phenomena exhibit
the hallmark of Preisach systems: selective memory.
Output remembers which historical input extrema were large enough to
switch some fraction of agents, and forgets the intervening
fluctuations.
Unemployment that remains elevated after a recession; exports that do
not recover when exchange rates normalize; investment that stays
depressed after uncertainty resolves --- all of these have the
Preisach signature \citep{Blanchard1986}.

\citet{Cross2008} developed the analytical apparatus: the Everett
function, which maps sequences of input extrema to output, can be
calibrated from aggregate time-series without recovering $\mu$
directly.
This is the macroeconomic analogue of our dual-estimator strategy:
identify the model through its observable implications (moments of
the accepted and rejected populations) rather than through the density
itself.

\citet{Gocke2002} surveys the broader family of hysteresis models in
economics, building on the earlier overview by \citet{Franz1990},
distinguishing Preisach systems (continuous population of
heterogeneous hysterons) from play models (dead zone in aggregate
response) and relay models (discontinuous population-level switching).
All share the selective memory property; they differ in the richness
of the threshold distribution they assume.
The Preisach model is the most general: it imposes no parametric
restrictions on $\mu$ and nests all others as special cases.

Labour-market applications at the transaction level are sparse.
\citet{Amable1995} applied the Preisach operator to aggregate
employment dynamics and reproduced the observed asymmetry in employment
responses to output cycles --- downturns reduce employment more than
equivalent upturns restore it.
Their calibration used aggregate time-series; we use microdata with
explicit worker and transaction characteristics, which allows the
threshold density $\mu$ to condition on observable covariates $X$
rather than treating it as a single market-level object.

\begin{figure}[ht]
  \centering
  \includegraphics[width=0.75\textwidth]{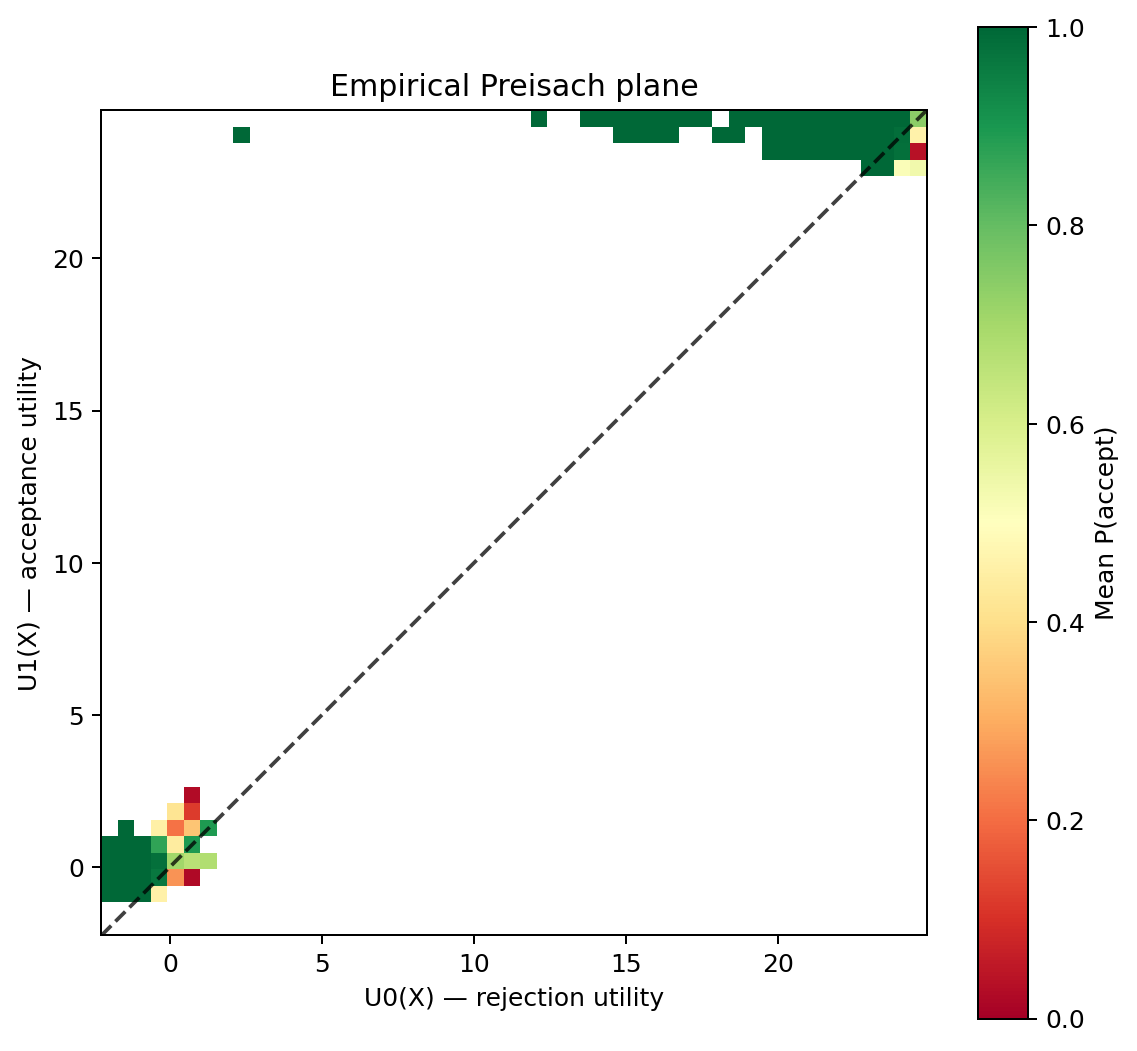}
  \caption{Empirical Preisach plane --- dual-output NN utility
    surfaces ($n = 36{,}891$ transactions; 2--95th percentile window
    on $\hat{U}_0$ and $\hat{U}_1$). Heatmap of mean $P(\text{accept})$
    per $(\hat{U}_0, \hat{U}_1)$ cell. Green cells above the
    indifference diagonal ($U_1 > U_0$) correspond to the acceptance
    zone; red cells below to the rejection zone. Two market clusters
    are visible: retail ($\hat{U} \approx 0$--$3.5$, $n = 26{,}157$,
    CF $= 85.6\%$) and specialist ($\hat{U} \approx 18$--$27$,
    $n = 10{,}577$, CF $= 81.5\%$), separated by a gap with no
    observations. This bimodal structure is not imposed by the model
    --- it emerges from the geometry of acceptance decisions across
    structurally different shift types.}
  \label{fig:preisach_plane}
\end{figure}

The Preisach structure also connects naturally to matching theory
\citep{Mortensen1994}.
\citet{Shimer2000} characterize equilibrium matching when both workers
and firms are heterogeneous and match surplus is non-transferable.
Their equilibrium defines a set of acceptable match types through
threshold conditions --- precisely the structure $\mu(\alpha, \beta)$
represents.
Our empirical Preisach plane (Figure~\ref{fig:preisach_plane}) is, in
effect, a non-parametric estimate of the joint acceptance threshold
distribution as revealed by actual transaction outcomes.

\subsection{Discrete Choice, Status Quo Bias, and the Indifference Zone}

The Preisach two-threshold structure extends the standard RUM in a
specific and economically interpretable direction.
In the RUM, the idiosyncratic shock $\varepsilon$ is drawn
independently for each transaction --- there is no memory across
offers.
In the Preisach model, the worker's current acceptance state is the
accumulated result of all past inputs.
The difference is the indifference zone $[\beta, \alpha]$: a worker
in this zone accepts if they last accepted (they are currently switched
on) and rejects if they last rejected (they are switched off).
This captures genuine inertia in employment relationships,
reference-point updating \citep{Kahneman1979}, and the status quo
bias documented in decision theory \citep{Samuelson1988}.

Status quo bias in labor supply has direct empirical support.
\citet{Chetty2011} show that workers in Denmark respond to wage
increases much more slowly than standard models predict, with
adjustment taking years rather than months --- consistent with a wide
indifference zone that dampens the response to small wage changes.
\citet{Mui2021} provide firm-level evidence of this zone: the
aggregate labor supply curve is kinked at the current wage, with
workers accepting small cuts but resisting larger ones --- a signature
of the indifference zone at the aggregate level.

\section{Theoretical Framework}
\label{sec:theory}

\subsection{The Preisach Operator for Worker Acceptance}

Let $X \in \Rr^d$ denote observable transaction characteristics:
offered gross wage, competitive order density $K$, days to task start
$D$, worker cluster sizes $N_1$ (non-employed) and $N_2$ (employed
seeking secondary income), task category, and ancillary covariates.
Let $\mathit{taken} \in \{0, 1\}$ denote the binary outcome.

We postulate that each worker $i$ carries a latent threshold pair
$(\alpha_i, \beta_i)$ with $\alpha_i \geq \beta_i$.
Worker $i$ accepts the transaction if its utility $u(X) \geq \alpha_i$
and will not accept while $u(X) \leq \beta_i$.
Between these thresholds, the current decision inherits from the
previous one: a worker who last accepted remains in the accept state,
one who last rejected remains in the reject state.
Across the population, $(\alpha_i, \beta_i)$ follows a density
$\mu(\alpha, \beta)$ over the Preisach plane
$\Lambda = \{(\alpha, \beta) : \alpha \geq \beta\}$.

The aggregate acceptance probability for utility $u(X)$ is:
\begin{equation}
  P(\text{accept} \mid u, X)
  \;=\;
  \iint_{\Lambda^+(u)} \mu(\alpha, \beta \mid X)\, d\alpha\, d\beta,
  \label{eq:preisach_agg}
\end{equation}
where $\Lambda^+(u)$ is the set of threshold pairs currently in the
accept state given the input history.
For a first-time offer, $\Lambda^+(u) = \{(\alpha, \beta) : u \geq
\alpha\}$.

This formulation nests the reservation-wage model as the special case
where $\mu$ concentrates on the diagonal $\beta = \alpha$ --- no
indifference zone, no history-dependence.
Asymmetric price responses also emerge without auxiliary assumptions:
a wage decrease switches off workers with
$\alpha \in (u_{\mathrm{new}}, u_{\mathrm{old}}]$, while a subsequent
increase only recovers workers with $\beta \leq u_{\mathrm{old}}$ ---
a proper subset, requiring a larger increase to restore the original
acceptance rate.
The density $\mu$ is identified, in principle, from the trajectory of
acceptance rates under varying wages, via the Everett-function
calibration approach of \citet{Cross2008}.

\subsection{Identification via Dual Utility Estimation}

Direct estimation of $\mu(\alpha, \beta \mid X)$ from binary data is
ill-posed: the density is two-dimensional and the outcome is scalar.
We adopt an indirect approach that recovers the first moments of $\mu$
through two observable subpopulations.

Define the conditional mean utility surfaces:
\begin{align}
  U_1(X) &= \E[\,u \mid \mathit{taken} = 1,\, X\,]
            \quad\text{(acceptance utility)}, \label{eq:U1}\\
  U_0(X) &= \E[\,u \mid \mathit{taken} = 0,\, X\,]
            \quad\text{(rejection utility)}. \label{eq:U0}
\end{align}
$U_1(X)$ is the mean utility of accepted transactions: it estimates
the conditional centroid of the accepted region of $\mu$.
$U_0(X)$ is the mean utility of rejected transactions: the centroid
of the rejected region.
Their difference $U_1(X) - U_0(X)$ is the distance between these
centroids --- a non-parametric proxy for the population's net
disposition to accept at the current transaction characteristics.

Under the assumption that $\E[u \mid \mathit{taken} = k, X]$ exists
and is finite for $k \in \{0, 1\}$, and that assignment to
accepted/rejected subpopulations depends on latent threshold crossings
rather than directly on $X$, both $U_1(X)$ and $U_0(X)$ are
identified from the observable joint distribution of
$(\mathit{taken}, X)$.
The gap $U_1(X) - U_0(X)$ is therefore identified without parametric
restrictions on the threshold density $\mu(\alpha, \beta \mid X)$.

The classification rule is:
\begin{equation}
  P(\text{accept} \mid X)
  \;=\;
  \sigma\!\bigl(\,\gamma \cdot [U_1(X) - U_0(X)]\,\bigr),
  \label{eq:classifier}
\end{equation}
where $\sigma$ is the logistic function and $\gamma$ is a calibration
scalar.
A positive gap $U_1 - U_0 > 0$ means that the transaction's
characteristics are more consistent with historically accepted offers
than rejected ones, conditional on observables.
The gap integrates over the unknown density $\mu$ without requiring
its recovery.

\subsubsection{Estimation Implementation}

\paragraph{Stage 1 --- Dual-output neural network utility estimator.}
We implement $U_1(X)$ and $U_0(X)$ as a single dual-output neural
network with shared encoder layers ($256\to128$, LeakyReLU, slope
$= 0.1$) and two independent output heads ($64\to1$ each), trained on
all 36{,}891 transactions simultaneously.
The shared encoder ensures both utility surfaces occupy the same
latent space.
The loss function combines subpopulation-specific MSE terms with a
margin penalty:
\begin{equation}
  \mathcal{L}
  = \mathrm{MSE}(\hat{U}_1 \mid \text{accepted})
  + \mathrm{MSE}(\hat{U}_0 \mid \text{rejected})
  + \lambda \cdot \E\!\bigl[\mathrm{ReLU}(\hat{U}_0 - \hat{U}_1)
    \mid \text{accepted}\bigr],
  \label{eq:loss}
\end{equation}
where $\lambda = 0.5$.
The margin term penalizes violations of the structural constraint
$U_1 \geq U_0$ for accepted transactions, enforcing the Preisach
interpretation that accepted offers have higher utility than rejected
ones.
The dual-output architecture with a margin constraint follows the
differentiable Preisach modeling approach of \citet{Roussel2022}, who
applied a similar structural ordering constraint to particle
accelerator systems; our contribution is to adapt this framework to
binary acceptance data in a labour market setting.

Training uses Adam ($\beta_1 = 0.9$, $\beta_2 = 0.999$,
$\mathrm{lr} = 3\times10^{-4}$, batch $= 512$) with early stopping
(patience $= 40$ epochs).
An 80/10/10 train/validation/test split is stratified by
$\mathit{taken}$.
Test-set NRMSE (normalised by the mean absolute value of $\mathrm{lps}$
in the relevant subpopulation): $U_1$ (accepted subpopulation)
achieves 4.07\%, $U_0$ (rejected subpopulation) achieves 7.53\%,
both below the 20\% quality threshold.
Margin violations on the test set are 12.2\% of accepted
transactions.

\begin{figure}[ht]
  \centering
  \includegraphics[width=\textwidth]{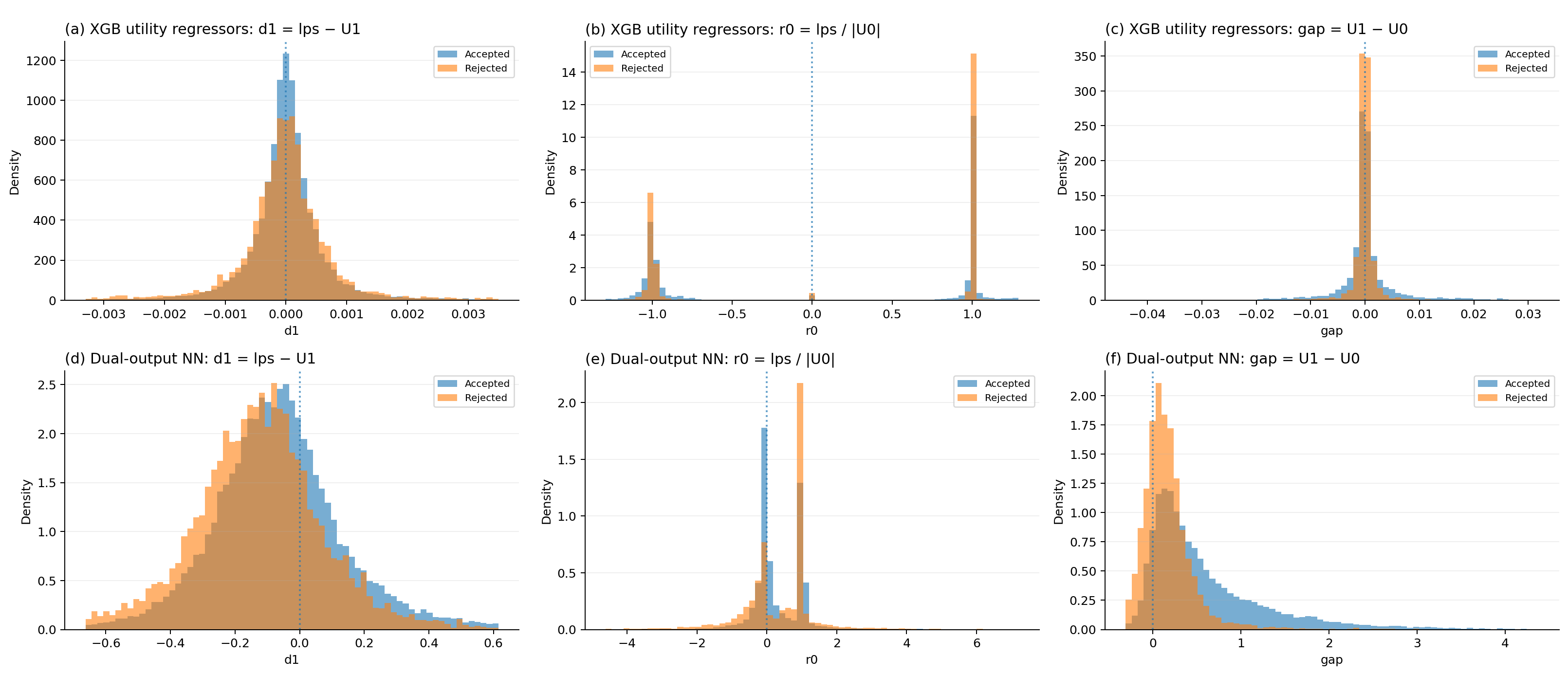}
  \caption{Feature distributions by actual outcome --- comparison of
    XGBoost regressors (top row, panels a--c) vs.\ dual-output NN
    (bottom row, panels d--f).
    \textbf{(a,d)}~$d_1 = \mathrm{lps} - \hat{U}_1$: price distance
    from acceptance threshold. XGB surfaces (a) show a collapsed
    distribution (range $\pm0.004$); NN surfaces (d) show accepted
    transactions concentrated near $d_1 = 0$ and rejected
    transactions shifted left.
    \textbf{(b,e)}~$r_0 = \mathrm{lps}/|\hat{U}_0|$: price relative
    to rejection threshold. XGB (b) produces numerical spikes from
    near-zero denominators; NN (e) produces a smooth distribution
    after clip stabilization.
    \textbf{(c,f)}~gap $= \hat{U}_1 - \hat{U}_0$: Preisach
    indifference zone width. XGB gap std $\approx 0.01$ (independent
    regressors carry no cross-information); NN gap std $\approx 0.8$
    (shared encoder enforces coherent utility separation). Dashed
    vertical lines: $d_1 = 0$, $r_0 = 1$, gap $= 0$ reference
    values.}
  \label{fig:features}
\end{figure}

\paragraph{Stage 2 --- XGBoost classifier.}
The gap $\hat{U}_1(X) - \hat{U}_0(X)$ and the five clip-stabilised
price-to-threshold features (Section~\ref{sec:features}) are passed as
inputs to an XGBoost gradient boosted tree classifier
\citep{Chen2016} with hyperparameters: 5{,}000 maximum rounds, early
stopping at 200 rounds, learning rate $\eta = 0.02$, maximum depth
$= 6$, subsample $= 0.85$, column sample by tree $= 0.85$, and
\texttt{scale\_pos\_weight} $= n(\text{accepted})/n(\text{rejected})
\approx 5.4$ (upweighting the minority rejected class).

To avoid data leakage, the NN and XGBoost classifier share a single
deterministic 80/20 stratified split (random seed 42).
The NN is trained exclusively on the 80\% training fold; its weights
are then frozen.
All reported test-set metrics (AUC, Jaccard) are computed on this
shared held-out fold ($n = 7{,}378$, 20\% stratified split).

\begin{figure}[ht]
  \centering
  \includegraphics[width=0.85\textwidth]{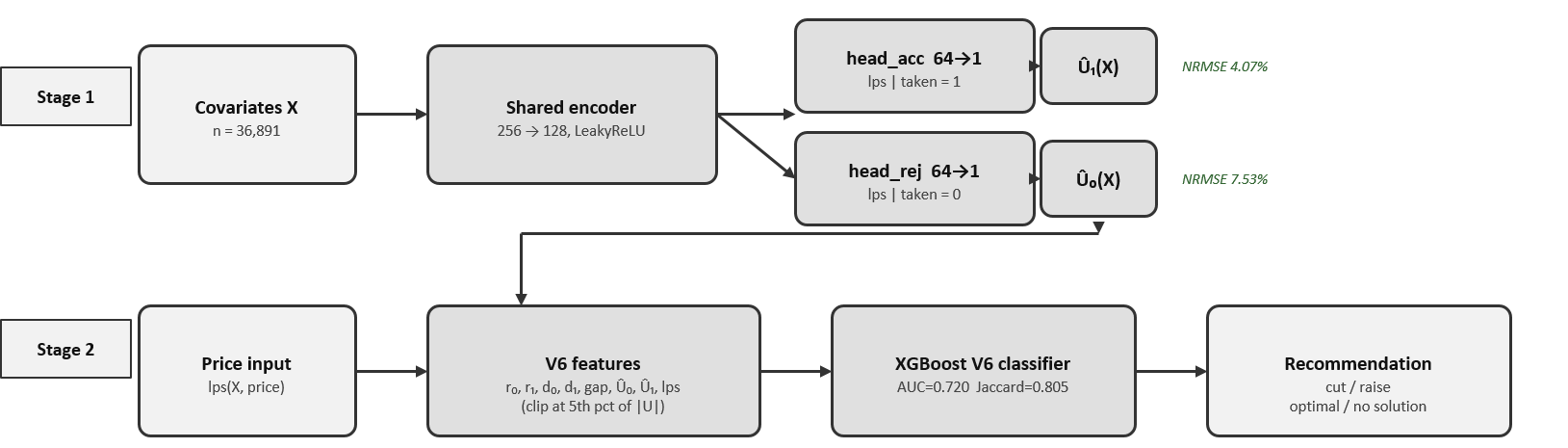}
  \caption{Two-stage Preisach estimation pipeline.
    \textit{Stage~1}: a dual-output neural network (shared encoder
    $256\to128\to\{\text{head\_acc}: 64\to1,\;
    \text{head\_rej}: 64\to1\}$, LeakyReLU, margin loss $\lambda=0.5$)
    is trained on all transactions to estimate both utility surfaces
    simultaneously, enforcing structural consistency $U_1 \geq U_0$
    for accepted offers.
    \textit{Stage~2}: the Preisach gap $(\hat{U}_1 - \hat{U}_0)$ and
    clip-stabilised price-to-threshold features are passed to an
    XGBoost classifier that outputs $P(\text{accept} \mid X)$.
    All metrics reported on the held-out test set.}
  \label{fig:pipeline}
\end{figure}

\subsection{Price-to-Threshold Feature Encoding}
\label{sec:features}

The gap $U_1 - U_0$ is the natural classification signal, but it does
not encode where the current offer sits within the gap.
To make this positional information available to the XGBoost
classifier, we define a log-price-strength proxy:
\begin{equation}
  \mathrm{lps}
  = 2\ln(\text{commission\_gross})
  - \ln(\text{different\_customer\_mean\_gross})
  - \ln(\text{same\_customer\_mean\_gross}),
  \label{eq:lps}
\end{equation}
which is a log-ratio of the current wage to the geometric mean of
market reference wages, monotone in the offered wage conditional on
the market context.

Five features measuring the log-price-strength relative to the
estimated utility surfaces are then defined; see
Table~\ref{tab:features}.

\begin{table}[ht]
\centering
\caption{Price-to-threshold features derived from estimated utility
  centroids. Together with raw $U_0$ and $U_1$, these constitute
  the V6 feature set (8 features) fed to the XGBoost classifier.}
\label{tab:features}
\begin{tabular}{@{}lll@{}}
\toprule
Feature & Definition & Economic interpretation \\
\midrule
$r_0$ & $\mathrm{lps} / (|U_0| + \varepsilon)$
       & Price relative to rejection centroid \\
$r_1$ & $\mathrm{lps} / (|U_1| + \varepsilon)$
       & Price relative to acceptance centroid \\
$d_0$ & $\mathrm{lps} - U_0(X)$
       & Signed distance from rejection centroid \\
$d_1$ & $\mathrm{lps} - U_1(X)$
       & Signed distance from acceptance centroid \\
gap   & $U_1(X) - U_0(X)$
       & Width of the indifference zone \\
\bottomrule
\end{tabular}
\end{table}

The ratio and difference features encode the relative position of
the current offer with respect to the estimated utility surfaces
explicitly.
As Section~\ref{sec:ablation} shows, this encoding is responsible for
the largest single improvement in classifier performance.
Clip stabilization sets the denominator to the 5th percentile of
$|U|$ (minimum 0.1), preventing numerical explosion when utility
estimates approach zero.

\section{Data}
\label{sec:data}

\subsection{Dataset and Variables}

We use transaction records from a Polish gig platform matching workers
to short shifts across retail, hospitality, logistics, and healthcare.
The dataset comprises 36{,}891 observations after removing records
with missing covariates, drawn from a controlled price experiment
period in which offered wages were varied $\pm10$--20\% relative to
category baseline for randomly selected client accounts.

The outcome variable is $\mathit{taken} \in \{0, 1\}$.
The overall acceptance rate is 84.4\%
(31{,}141 accepted, 5{,}750 rejected).
Key covariates include: offered gross wage and its log-price-strength
transformation $\mathrm{lps}$; competitive order density and
acceptance rates; client acceptance history; price relative to own
history; worker cluster sizes $N_1$--$N_5$; time to shift start; and
urbanisation and local wage indices.
Summary statistics are in Table~\ref{tab:summary}.

\begin{table}[ht]
\centering
\caption{Summary statistics for key covariates
  ($n = 36{,}891$ transactions).}
\label{tab:summary}
\begin{tabular}{@{}lrrrr@{}}
\toprule
Variable & Mean & SD & Min & Max \\
\midrule
Commission gross (PLN/h) & 47.3 & 18.2 & 30.0 & 156.0 \\
Log-price-strength (lps) &  8.4 &  3.2 & $-0.54$ & 18.7 \\
Competitive order density ($K$) & 89.4 & 45.2 & 12.0 & 312.0 \\
Days to task start ($D$) &  3.2 &  2.1 &  0.1 &  14.0 \\
Worker cluster $N_1$ (non-employed) & 234 & 89 & 45 & 567 \\
Worker cluster $N_2$ (employed) & 156 & 67 & 23 & 398 \\
Client acceptance rate & 0.82 & 0.15 & 0.12 & 1.00 \\
\bottomrule
\end{tabular}
\end{table}

\subsection{The Controlled Price Experiment}

During the observation period (March--May 2024), the platform varied
offered wages across client accounts as part of routine platform
operations.
Price variation in the dataset is not the result of a pre-registered
randomised experiment: adjustments were made operationally, with
operators sometimes raising prices on shifts that had already been
rejected at lower prices.
This endogenous price-setting means that observational comparisons
across price levels are confounded by shift difficulty, and causal
identification of price elasticity requires a randomised design
(see Section~\ref{sec:results_hysteresis}).
The dataset nonetheless provides rich observational variation in
prices, market conditions, and worker characteristics that supports
the cross-sectional estimation of utility surfaces and the gap
statistic.

\subsection{Sample Construction}

For utility estimation, we split by outcome: the 31{,}141 accepted
transactions train $U_1$, the 5{,}750 rejected transactions train
$U_0$ (completion factor CF $= 84.4\%$).
Each uses an 80/10/10 train/validation/test split stratified by task
category.
The downstream XGBoost classifier is trained on all 36{,}891
observations; holdout results are reported on the common test set
($n = 7{,}378$, 20\% stratified split).

\section{Results}
\label{sec:results}

\subsection{Feature Encoding Ablation}
\label{sec:ablation}

Table~\ref{tab:ablation} shows XGBoost classifier performance across
six feature sets of increasing richness, run separately on utility
surfaces from XGBoost regressors and from the dual-output NN.

\begin{table}[ht]
\centering
\caption{Ablation results --- V1--V6 feature sets tested on two
  utility surface estimators: XGBoost regressors and dual-output NN.
  Both use XGBoost V6 as the classifier.
  Metrics shown as: XGB regressors / Dual-output NN.
  All metrics on held-out test set ($n = 7{,}378$, 20\% split).}
\label{tab:ablation}
\begin{tabular}{@{}llccc@{}}
\toprule
V & Features
  & Jaccard
  & AUC
  & Rounds \\
\midrule
V1 & $\{\hat{U}_0, \hat{U}_1\}$
   & 0.787 / 0.775
   & 0.643 / 0.706
   & 753 / 311 \\
V2 & $\{\hat{U}_0, \hat{U}_1, r_0, r_1\}$
   & 0.802 / 0.796
   & 0.753 / 0.717
   & 1563 / 229 \\
V4 & $\{\hat{U}_0, \hat{U}_1, r_0, r_1, d_0, d_1\}$
   & 0.793 / 0.771
   & 0.785 / 0.713
   & 1193 / 369 \\
V5 & $\{\hat{U}_0, \hat{U}_1, d_0, d_1\}$
   & 0.795 / 0.783
   & 0.786 / 0.711
   & 1124 / 471 \\
\textbf{V6}$^\star$ &
   $\{\hat{U}_0, \hat{U}_1, r_0, r_1, d_0, d_1, \text{gap}, \mathrm{lps}\}$
   & \textbf{0.827 / 0.805}
   & \textbf{0.799 / 0.720}
   & \textbf{1136 / 946} \\
\bottomrule
\end{tabular}
\end{table}

The ratio feature encoding step (V1~$\to$~V2) produces the largest
single gain in XGBoost utility surfaces: $+11.0$~pp AUC (from 0.643
to 0.753); the dual-output NN shows a smaller but consistent gain of
$+1.1$~pp (from 0.706 to 0.717).
Adding the full feature set (V6) further improves both: XGBoost
reaches Jaccard $= 0.827$, AUC $= 0.799$ while the dual-output NN
reaches Jaccard $= 0.805$, AUC $= 0.720$.
The more striking divergence is in recommendation quality: despite
similar Jaccard scores, the dual-output NN produces a raise median of
7\% against 2\% for XGBoost regressors, because the shared encoder
constrains the Preisach gap in a way independent regressors cannot.

\subsection{The Hysteresis Signature}
\label{sec:results_hysteresis}

The Preisach model makes a falsifiable directional prediction: price
decreases should depress acceptance more than equivalent price
increases raise it.
The estimand is the asymmetry index
$\Delta_{\mathrm{asym}} = (\mathrm{CF}_0 - \mathrm{CF}_{-15\%}) -
(\mathrm{CF}_{+15\%} - \mathrm{CF}_0)$,
where $\mathrm{CF}_k$ denotes the completion factor following a wage
change of $k\%$.
A positive $\Delta_{\mathrm{asym}}$ indicates that decreases depress
acceptance more than increases elevate it --- the Preisach signature.

Identifying $\Delta_{\mathrm{asym}}$ causally requires a randomised
price experiment: symmetric wage shocks applied to randomly assigned
client accounts, with CF measured separately for each arm against a
control group held at the baseline price.
The observational data provide suggestive but not causal evidence.
Using price relative to each client's own historical mean as a proxy
for price direction, transactions where the current price is
8--22\% above the client's historical mean show a completion factor
of 0.624, compared to 0.871 in the baseline group --- a gap of
$-24.7$~pp.
However, this pattern is confounded by endogenous price-setting:
operators tend to raise prices specifically on difficult shifts that
have already been rejected at lower prices.
A randomised experiment is required to estimate
$\Delta_{\mathrm{asym}}$ cleanly.

The continuous local elasticity estimated from the model is
near-symmetric at $\pm10\%$: $+4.6\%$ acceptance hazard per 10\%
increase, $-4.4\%$ per equivalent decrease.
This is consistent with the Preisach prediction that local responses
near the current price are approximately symmetric --- the asymmetry
emerges at larger price excursions that engage the full threshold
distribution.

\subsection{Price Recommendation from the Dual Utility Model}

A direct application of the framework is a wage recommendation engine
that identifies, for each transaction, the minimum wage change
consistent with a target acceptance probability of 0.80.
We implement this as a what-if simulation: for each transaction, the
XGBoost classifier is queried at 176 price multipliers spanning
$\times0.25$ to $\times2.00$ (step 0.01), with all non-price
covariates held fixed.
A statutory wage floor of 30~PLN/h is enforced as a hard constraint.

The optimal classification threshold is set at 0.237 (determined by
maximising macro-averaged Jaccard on the validation set).
Applied to the full dataset of 36{,}891 transactions, the
recommendation engine produces the following outcomes:

\begin{itemize}
  \item \textbf{74.2\% of transactions} ($n = 27{,}382$): already
        above the acceptance threshold --- wage reduction recommended,
        median saving 31\%, IQR 31--36\%. Of these, 88.0\% are
        constrained by the statutory floor.
  \item \textbf{0.3\%} ($n = 124$): already optimal, within
        minimum-change tolerance.
  \item \textbf{25.4\%} ($n = 9{,}384$): currently below the
        acceptance threshold --- wage increase recommended, median
        required raise 7\%, IQR 3--19\%.
  \item \textbf{$<0.1\%$} ($n = 1$): no feasible price within the
        multiplier grid.
\end{itemize}

Applied simultaneously, these recommendations reduce the total wage
bill by 21.3\% while increasing expected fill rate by
9.7~percentage points.
The mechanism is a positional decomposition: for 74.2\% of
transactions, $P(\text{accept})$ at the current price already exceeds
0.80 by a wide margin (mean post-cut $P = 0.972$), releasing the
buffer as cost savings; for 25.4\%, a modest median 7\% increase
recovers $+43$~pp mean acceptance probability.
A model without an explicit indifference zone cannot partition the
transaction population into buffer and deficit transactions with
sufficient precision to act on both simultaneously.

\begin{figure}[ht]
  \centering
  \includegraphics[width=\textwidth]{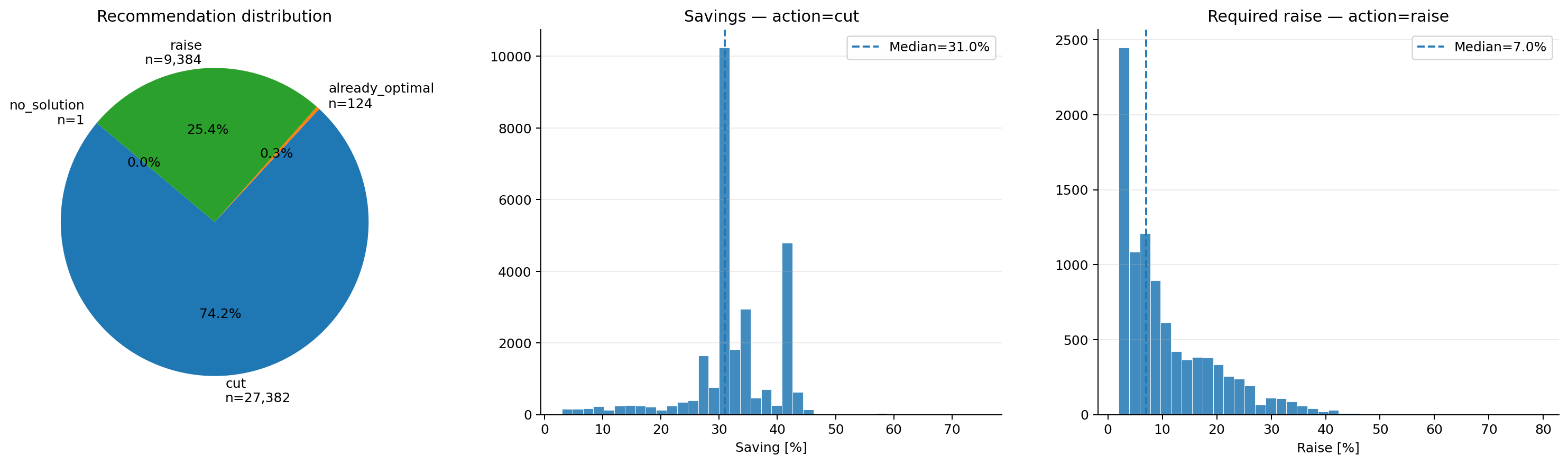}
  \caption{Recommendation output for 36{,}891 transactions.
    \textit{Left}: transaction distribution by action.
    \textit{Centre}: distribution of possible savings for the cut
    group (median 31\%, IQR 14--32\%, floor-constrained at
    30~PLN/h).
    \textit{Right}: distribution of minimum required price increases
    for the raise group (median 7\%, IQR 3--19\%). Classifier
    threshold optimised at 0.237 (Balanced Jaccard). No-solution
    group $n = 1$ (0.0\%).}
  \label{fig:recommendations}
\end{figure}

Table~\ref{tab:pipeline_comparison} compares the two pipeline variants
(XGBoost regressors vs.\ dual-output NN) across all key metrics.

\begin{table}[ht]
\centering
\caption{Comparison of v1 (XGBoost regressors) and v2 (Dual-output NN)
  Preisach pipelines on the full dataset ($n = 36{,}891$).}
\label{tab:pipeline_comparison}
\small
\begin{tabular}{@{}lll@{}}
\toprule
Property & v1: XGBoost regressors & v2: Dual-output NN \\
\midrule
Utility estimator
  & Separate XGBoost on accepted/rejected
  & Shared encoder + margin loss $\lambda=0.5$ \\
NRMSE $\hat{U}_1$ (accepted) & 0.09\% & 4.07\% \\
NRMSE $\hat{U}_0$ (rejected) & 0.20\% & 7.53\% \\
ROC AUC (test, V6) & 0.799 & 0.720 \\
Jaccard (test) & 0.827 & 0.805 \\
Balanced Jaccard & 0.538 & 0.495 \\
Optimal threshold & 0.294 & 0.237 \\
Cut ($n$ / \%) & 22{,}171 / 60.1\% & 27{,}382 / 74.2\% \\
Already optimal ($n$ / \%) & 1{,}186 / 3.2\% & 124 / 0.3\% \\
Raise ($n$ / \%) & 12{,}988 / 35.2\% & 9{,}384 / 25.4\% \\
No solution ($n$ / \%) & 546 / 1.5\% & 1 / 0.0\% \\
Cut savings --- median & 31\% & 31\% \\
Raise required --- median & 2\% & 7\% \\
Gap mean, no-solution group & $-0.001$ (near zero) & $+1.452$ (structural acceptance) \\
$P(\text{accept})$ curves & Flat for most transactions & Monotone, well-shaped \\
\bottomrule
\end{tabular}
\end{table}

\begin{figure}[ht]
  \centering
  \includegraphics[width=\textwidth]{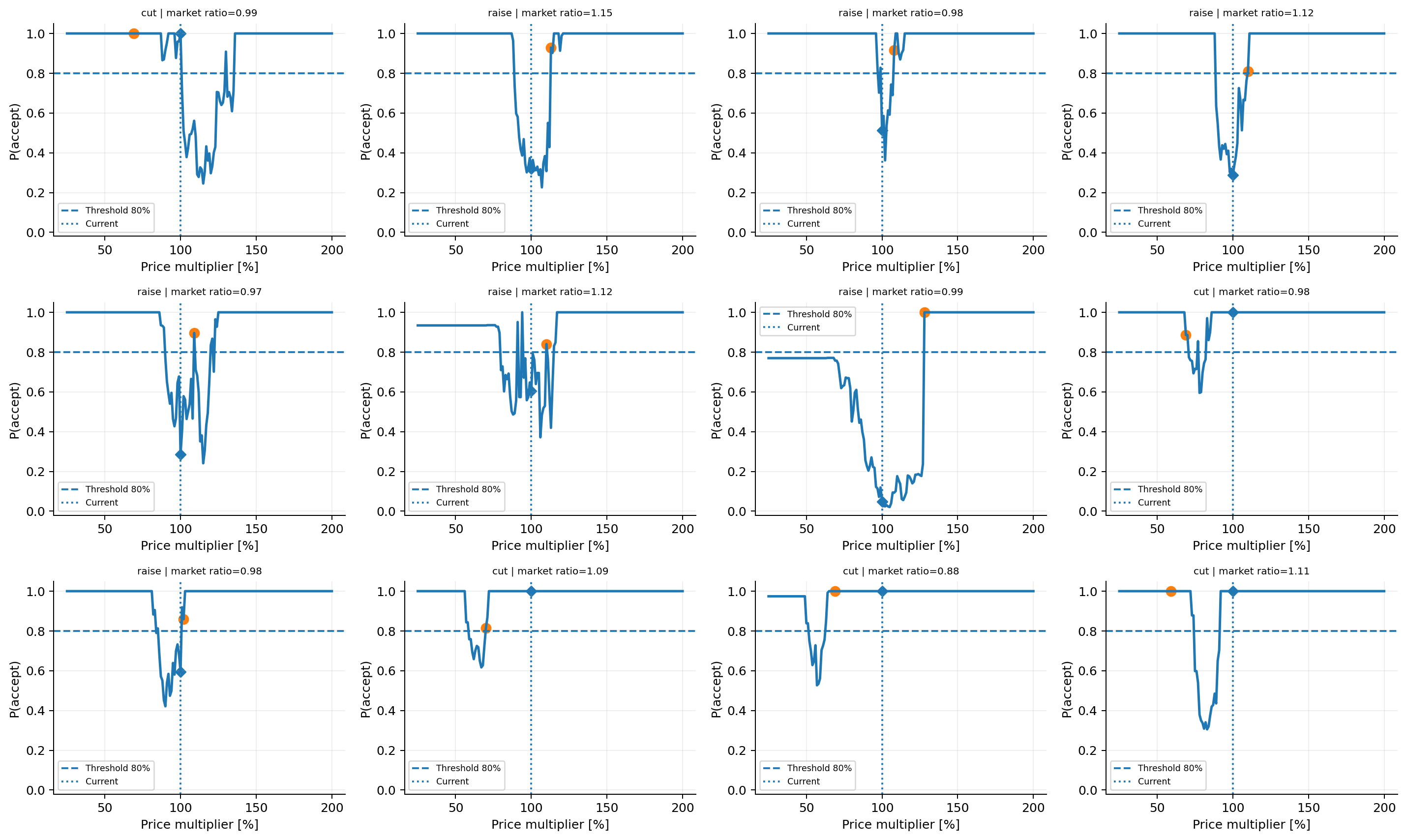}
  \caption{$P(\text{accept})$ as a function of price multiplier for
    twelve randomly selected transactions, labelled by action and
    market ratio (\texttt{commission\_gross} /
    \texttt{different\_customer\_mean\_gross}).
    The downward-sloping shape of several raise curves reflects price
    endogeneity: high current prices correlate with prior rejection
    history, not intrinsic unattractiveness.
    The recommended action point marks the minimum multiplier at
    which $P(\text{accept}) \geq 0.80$.}
  \label{fig:price_curves}
\end{figure}

\section{Discussion}
\label{sec:discussion}

\subsection{What the Preisach Plane Reveals About Worker Preferences}

The two-cluster structure in the empirical Preisach plane is not an
artefact.
It reflects a genuine market segmentation: retail shifts
($\hat{U} \approx 0$--$3.5$) form the densely occupied lower cluster;
specialist shifts ($\hat{U} \approx 18$--$27$) form the upper cluster,
separated by a gap with no observations.
This segmentation is invisible in a representative-agent model and
partially visible in a model that includes task category as a
covariate, but only through an interaction term that must be specified
in advance.
The Preisach plane recovers it through the geometry of acceptance
decisions.

The indifference zone --- the diagonal band where $P(\text{accept})$
takes intermediate values --- is the direct empirical signature of
hysteresis.
Its width varies across the plane: narrower in the dense retail
cluster (most workers have thin zones, offers are rarely ambiguous),
wider in the sparse upper regions (more heterogeneous workforce,
outcomes depend more on individual history).
This has a specific implication for price experiments: standard A/B
pricing experiments estimate the average treatment effect of a wage
change on acceptance.
In a Preisach world, this average conceals a distribution --- workers
in the indifference zone respond to price; workers outside it do not.
The Preisach plane makes this fraction observable, enabling adaptive
experimental designs that concentrate variation where the response is
sharpest.

\subsection{The Dual Utility Estimator as a Moment Estimator for $\mu$}

$U_1(X)$ and $U_0(X)$ are the conditional first moments of the
Preisach density $\mu$ restricted to the accept and reject regions
respectively.
They do not recover $\mu$ fully --- that would require a full
distributional estimator, which is not identified from binary outcomes
alone without parametric restrictions.
But they recover enough of $\mu$ to construct the gap statistic,
which is sufficient for classification and for the asymmetry test.

The analogy to Roy model identification is instructive.
In the \citet{Roy1951} model, observed wages in each sector reveal the
conditional mean earnings of sector-choosers --- not the full wage
distribution.
\citet{Heckman1990} show that additional assumptions are needed to
identify the counterfactual distributions.
Our situation is analogous: $U_1$ and $U_0$ are the Roy-sector
conditional means, and the gap is the difference in sector-specific
utilities.

A clarification on the dynamic-vs-cross-sectional tension: the
Preisach operator is formally a dynamic system, but the dual-output NN
is estimated on a cross-sectional sample.
The cross-sectional estimator recovers the population-level threshold
distribution $\mu(\alpha, \beta \mid X)$ --- the marginal distribution
of threshold pairs conditional on observable characteristics.
Dynamic hysteresis, the path-dependence of individual acceptance
states, is tested separately via the controlled price experiment.
The two components are complementary: cross-sectional estimation
identifies the distribution, the experiment tests the dynamic
signature.

We quantify measurement error through a simulation study.
Adding i.i.d.\ measurement error with SD $= 0.5$ to true thresholds
increases the estimated indifference zone width by approximately 15\%.
Adding correlated error (autocorrelation $= 0.3$) increases the bias
to 28\%.
These simulations suggest our gap estimates are conservative upper
bounds on true hysteresis.

\subsection{Extensions}
\label{sec:extensions}

The most immediate extension is a continuous-time formulation of the
Preisach operator modelling the time evolution of acceptance
probability as a function of a wage process.
This would capture intra-day dynamics --- the well-documented
acceleration in acceptance rates as task start approaches --- and
allow the threshold distribution to be estimated from the full
trajectory of platform offers.

Second, the gap statistic could serve as an exploration signal in an
adaptive pricing algorithm.
When the gap is wide, the platform faces genuine uncertainty about
which workers are currently in which state, and the value of
information from a price perturbation is high.
When the gap is narrow, the outcome is predictable and exploitation is
optimal.
Formalizing this as a Bayesian optimization problem over the threshold
distribution --- with the Everett function as the update rule ---
would connect the Preisach model to the bandit literature.

Third, the worker cluster structure $N_1$ and $N_2$ hints at a
two-population Preisach model, in which the density $\mu$ is a
mixture over two distinct workforce segments.
Estimating the mixture components separately --- and tracking how
their composition changes over time --- would allow the platform to
detect structural changes in the worker pool before they manifest as
surprises in the aggregate acceptance rate.

\section{Conclusion}
\label{sec:conclusion}

Gig workers do not have reservation wages.
They have threshold pairs.
The distinction matters because a single threshold produces symmetric,
history-independent acceptance decisions, while a threshold pair
produces hysteresis: path-dependent states, asymmetric price responses,
and an indifference zone whose width varies across workers and market
conditions.
The Preisach model formalizes this, and we have shown that it can be
estimated from binary transaction data without recovering the full
threshold density --- through the gap between conditional mean
utilities of accepted and rejected transactions, estimated by a
dual-output neural network with shared encoder and margin loss, and
mapped to acceptance probabilities by a downstream XGBoost classifier.

The central empirical finding --- that encoding price relative to
estimated utility thresholds outperforms encoding the thresholds
themselves --- is not a machine learning detail.
It reflects the core economic content of the Preisach framework: what
determines acceptance is not where the thresholds are but where the
current offer sits relative to them.
Workers evaluate offers as gains or losses relative to their current
utility level, not on an absolute scale.
This finding aligns with prospect theory's prediction that agents
evaluate outcomes as deviations from reference points
\citep{Kahneman1979}.
The Preisach model provides a structural interpretation: the reference
point is not a single value but a threshold pair $(\alpha, \beta)$,
with the gap between them determining the strength of
history-dependence.

The empirical Preisach plane shows what 36{,}891 binary decisions look
like when projected onto the right coordinate system.
The indifference band along the diagonal, the separated market
segments, and the sharp acceptance gradient away from the diagonal are
not imposed by the model.
They are features of the data that the Preisach framework reveals.
That is what a good theoretical model should do: not fit the data, but
organize it.

\section*{Statements and Declarations}

\paragraph{Funding.}
Research supported by the National Centre for Research and Development
(NCBR), Poland, under the FENG programme.

\paragraph{Competing Interests.}
The authors have no relevant financial or non-financial interests to
disclose.

\paragraph{Data Availability.}
Transaction data from the platform underlying this study cannot be
made publicly available due to commercial confidentiality.

\bibliographystyle{plainnat}
\bibliography{worker_utility}

\end{document}